\documentclass[10pt,twocolumn,letterpaper]{article}

\usepackage{cvpr}              

\usepackage{graphicx}
\usepackage{amsmath}
\usepackage{amssymb}
\usepackage{booktabs}

\usepackage{todonotes}
\usepackage{siunitx}

\usepackage[pagebackref,breaklinks,colorlinks]{hyperref}

\usepackage[capitalize]{cleveref}
\crefname{section}{Sec.}{Secs.}
\Crefname{section}{Section}{Sections}
\Crefname{table}{Table}{Tables}
\crefname{table}{Tab.}{Tabs.}

\def\cvprPaperID{45} 
\def\confName{CVPR}
\def\confYear{2023}

\usetikzlibrary{calc}

\begin{document}

\title{Interpolation-Based Event Visual Data Filtering Algorithms}

\author{Marcin Kowalczyk, Tomasz Kryjak\\
Embedded Vision Systems Group, AGH University of Krakow\\
{\tt\small kowalczyk@agh.edu.pl} {\tt\small tomasz.kryjak@agh.edu.pl}
}
\maketitle

\tikz[overlay,remember picture] {
	\node at ($(current page.north)+(0,-1.5)$) {\textcolor{gray}{This paper has been accepted for publication at the}};
	\node at ($(current page.north)+(0,-2.0)$) {\textcolor{gray}{IEEE Conference on Computer Vision and Pattern Recognition (CVPR) Workshops, Vancouver, 2023. ©IEEE}};
}

\begin{abstract}

The field of neuromorphic vision is developing rapidly, and event cameras are finding their way into more and more applications.
However, the data stream from these sensors is characterised by significant noise.
In this paper, we propose a method for event data  that is capable of removing approximately 99\% of noise while preserving the majority of the valid signal.
We have proposed four algorithms based on the matrix of infinite impulse response (IIR) filters method.
We compared them on several event datasets that were further modified by adding artificially generated noise and noise recorded with dynamic vision sensor.
The proposed methods use about 30KB of memory for a sensor with a resolution of 1280 $\times$ 720 and is therefore well suited for implementation in embedded devices.
\end{abstract}

\section{Introduction}
\label{sec:intro}

The field of computer vision has been developing rapidly for a long time. 
One of the areas that has become particularly popular in recent years is neuromorphic vision.
Dynamic vision sensors respond to changes in pixel brightness, rather than registering the absolute value of the intensity and colour of the light falling on the matrix pixels.
Cameras operating in this way have a number of advantages over traditional solutions (hereafter referred to as frame cameras).
The first is high temporal resolution.
Changes are monitored at a frequency of up to \SI{1}{\MHz} (depending on the sensor model).
This makes it possible to monitor very fast movements with very small motion blur (depends on the illumination of the observed scene \cite{hu2021v2e}), which is a common problem with frame cameras.
The second benefit is low latency. Thanks to the independent operation of the pixels, there is no need to wait for the entire image to be processed.
This allows to achieve real-world latencies in the order of about one to a few \si{\ms} \cite{hu2021v2e}.
The third is low power consumption, as only brightness change information is transmitted, eliminating data redundancy.
The power consumption of the sensor itself is up to a few hundred \si{\mW} \cite{gallego2020event}.
A final advantage is the high dynamic range. In the case of event cameras, this can even exceed \SI{120}{\dB}. This is considerably more than even high-quality conventional cameras, which have a dynamic range of around \SI{60}{\dB}. This allows operation in both very dark and very bright environments (even within the same scene).



\begin{figure}[!t]
    \centering
    \includegraphics[width=3in]{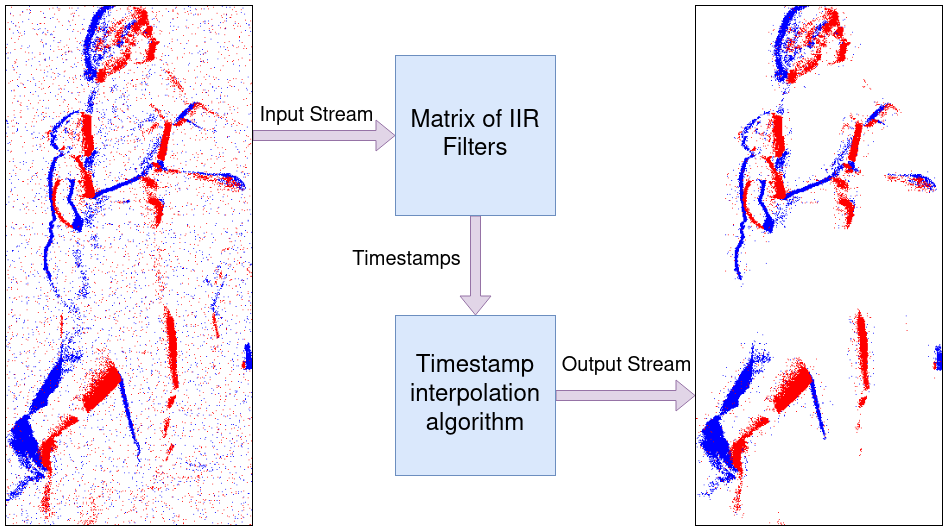}
    \caption{
    A simplified diagram of the proposed method and an example of input and output data, aggregated for \SI{20}{\ms}.
    }
    \label{fig::introduction}
\end{figure}

The use of event cameras also presents challenges with a new type of vision information.
The data obtained is a sparse point cloud in three-dimensional space (pixel coordinates, time of occurrence and polarity).
Therefore, well-known image processing algorithms developed over several decades cannot be directly applied to such data.
Information about the intensity and colour of the light falling on a given pixel is also not available. Only information about the increase or decrease in brightness of a given receptor on the matrix is obtained.
Due to the non-ideality of the sensors, photon shot noise and noise from the analogue part of the pixel array, these sensors are very prone to noise.
These appear as events that are not related to the movement of the scene or changes in illumination. 
They resemble impulse noise, which is very easy to see in the output data.
It is worth noting that the presence of noise in the processed data is a potential source of reduced algorithm performance.

Figure \ref{fig::introduction} shows a simplified schematic of the proposed method, with example input and output data. The events have been accumulated over a period of \SI{20}{\ms}. A large reduction in non-motion related noise is evident.

In this paper we propose 4 algorithms for event data filtering based on the solution presented in the paper \cite{kowalczyk2022hardware}.
The aim of the research described in this article was to develop an algorithm that requires a small amount of memory and at the same time has a high interference filtering efficiency, with as few correct events removed as possible. In order to reduce the number of valid events removed, the proposed algorithms should reduce the negative effects associated with the transition of observed objects between neighbouring regions, which was present in the original algorithm.
This type of solution is well suited for implementation in systems with a small amount of memory, or when excessive memory usage negatively affects performance (e.g. the processor is equipped with small cache resources).
The proposed methods have a high filtering efficiency for uncorrelated interference, comparable to popular filtering algorithms, such as the nearest neighbour method (NNb).



The rest of the article is structured as follows.
Chapter \ref{sec::previous} describes the research published so far on event sensor data filtering.
Section \ref{sec::basic_algorithm} presents the basic filtering algorithm developed in previous research.
The proposed filtering methods are then discussed in section \ref{sec::interpolation}.
Their evaluation and comparison is presented in section \ref{sec::evaluation}.
The article concludes with a summary and possible further development of the ongoing research in section \ref{sec::conclusion}.

\section{Previous work}
\label{sec::previous}
The topic of video filtering of event data has been addressed many times in the scientific literature due to its great practical importance. 
This subsection presents an analysis and comparison of different approaches to this problem.

A 2008 paper \cite{delbruck2008frame} presented one of the first event stream filtering methods.
It consisted of removing events for which no activity was recorded in their surroundings within a given time window. This allowed the removal of single, isolated events.
The requirement was to store a timestamp map with a size equal to the sensor resolution multiplied by 2. In this paper, a sensor with a resolution of 128 $\times$ 128 was used.

The filtering of event data with greyscale information is presented in the article \cite{ieng2014asynchronous}.
An asynchronous time-based image sensor (ATIS) with a resolution of 304 $\times$ 240 was used. Such sensors report events when the brightness of the pixels changes, but when an event is detected for a particular pixel, a reading of the incident light intensity is initiated. As a result, the greyscale value of the reporting pixel is also available for each event.
Asynchronous linear and non-linear filtering techniques have been proposed. 
These include the Gaussian blur filter, the bilateral filter and the Canny edge detector.

A framework for removing uncorrelated noise on an FPGA platform was presented in the paper \cite{linares2015usb3}. Timestamps corresponding to the processed events were stored in a register matrix.
For each input event, the previously stored timestamp was read from the matrix. The difference between the input one and the read one was then calculated. If this was greater than a preset threshold, the event was erased. A new timestamp was written to the same location in the array. In the work in question, this had a size of 128 $\times$ 128. The work also considered writing the received timestamp to neighbouring pixels.

In the paper \cite{liu2015design}, a background activity filter was proposed to pass on space-time correlated events for further processing.
The aim was to reduce communication and computational overhead while increasing the rate of correct information content. A 128 $\times$ 128 matrix chip with 20 $\times$ 20 \(\si{\um}^2\) cells was designed. 
Each cell combines subsampling in the spatial domain with a time window based on current integration.

The paper \cite{czech2016evaluating} compares eight filtering algorithms.
Three methods calculate the difference in timestamps between the pixel being processed and the last event in its vicinity.
The fourth requires that at least two other events belong to the neighbourhood of the processed event within the given time window. Otherwise it will be deleted.
The next one deletes the data if its polarity is the same as the previous event reported by the pixel.
The next two methods erase the data if not enough time has elapsed since the previous event reported by the pixel in question.
The last method calculates the average time between events reported by all pixels.

The article \cite{barrios2018less} introduces the LDSI algorithm (Less Data Same Information).
Its aim is to reduce the amount of data processed without removing relevant information.
It used impulse cells, inspired by the action of biological neurons, to process data in four layers.
Successive events increased the potential of the corresponding neurons and their neighbours in the third layer.

In the paper \cite{khodamoradi2018n}, a spatiotemporal filter with O(N) memory complexity was presented. In this method, instead of using one memory cell for each pixel, it was proposed to use two cells for each row and column, which allowed a significant reduction in memory usage. An implementation in an FPGA chip is also presented.

An implementation of a filtering algorithm using a pulse neural network in the IBM TrueNorth Neurosynaptic System neuromorphic processor is presented in the paper \cite{padala2018noise}.
A variation of the integrate-and-fire neuron model was used.
The network has two layers. 
The first introduces a break period. It reduces the maximum response frequency of the neurons to remove high frequency noise from a single pixel.
The second layer is a neural implementation of the NNb method. It checks whether other events have been generated in the vicinity of the event being processed.

In the paper \cite{bisulco2020near}, a method for event stream filtering and compression based on two switched time windows was presented.
These windows operate alternately in two phases.
In the first, events are input into one window and the data from the other window are processed and cleaned. In this way, an image representing the sensor data is created.
In the second phase, the operation of the windows is reversed. The data from the first window is read and cleaned, and the data is written to the second window.
However, as a result of the algorithm, the time stamps of individual events are removed and are not passed to the filter output.

A pedestrian detection system is presented in \cite{ojeda2020device}.
In its first part, a point process filter is proposed to filter out noise in the event data. It uses an adaptive time window. It allowed to increase the classification accuracy. In the second part, a binary neural network was used for classification.
The proposed filtering method outputs a two-dimensional image in which the recorded events are marked. They are aggregated within a given time window (\SI{3}{\ms}). In the generated image, the logical product of adjacent pixels is performed (separately in the horizontal and vertical directions).

The authors of the paper \cite{guo2020hashheat} proposed a hashing-based filtering algorithm that does not require the storage of 32-bit timestamps, thus achieving low memory requirements.
It was also implemented on an FPGA chip.

An impulse neural network was also presented in the paper \cite{xiao2021snn}.
It had an adaptive time window length. A leaky integrate-and-fire (LIF) neuron model was used.
According to the authors, the proposed method performs better than the classical filtering method with a time window.

Filtering inspired by the action of the human retina has been proposed in \cite{gupta2021foveal}.
A retinal model based on Gaussian difference filters inspired by the fovea of the optic nerve was used.
A sensor with a resolution of 128 $\times$ 128 was used in this work.

An adaptive background activity filtering method for event data was presented in \cite{mohamed2022dba}.
The authors used an optical flow and a non-parametric KNN (K-nearest neighbour) regression algorithm.
In the first stage of the algorithm, redundant events are removed using a technique based on a dynamic timestamp. In the second stage, background activity noise is removed using the adaptive KNN algorithm.

The paper \cite{guo2022low} proposes 3 algorithms for filtering event data.
The first calculates the distance to previous events stored in the first-in-first-out memory.
The second checks whether a certain number of other events have been registered in the vicinity of the processed event within a given time window.
The last is based on a simple neural network that processes the age of events from the analysed neighbourhood.

The literature review confirms the importance and popularity of event data filtering. Newer and newer algorithms are proposed to remove distortions.


\section{Basic version of the algorithm}
\label{sec::basic_algorithm}

In the paper \cite{kowalczyk2022hardware}, a filtering method based on an infinite impulse response filter array was proposed.
The proposed filtering method is able to remove more than 99\% of the uncorrelated noise. The operation of the algorithm is based on dividing the event camera matrix into regions. Each event is passed to a filter corresponding to the area of the event being processed. Each domain contains a timestamp. An event is removed if the difference between the internal area marker and the timestamp of the processed event is greater than a preset threshold.
The filtering algorithms proposed in this paper are an extension of this method.

\begin{figure}
    \centering
    \subfloat[]{\includegraphics[height=1.5in]{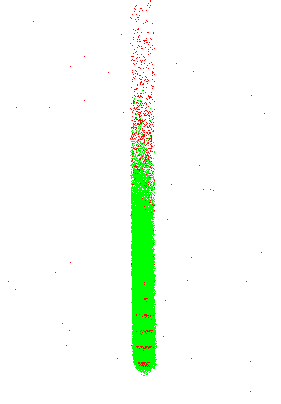}\label{sfig::corn_all_events}}
    \subfloat[]{\includegraphics[height=1.5in]{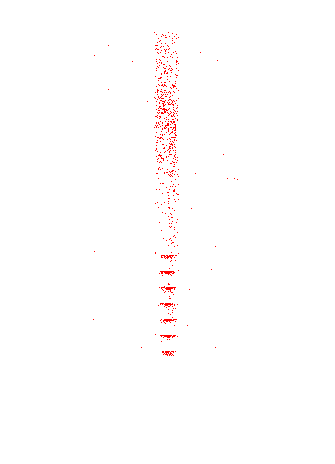}\label{sfig::corn_false_events_a}}
    \subfloat[]{\includegraphics[height=1.5in]{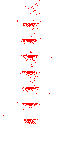}\label{sfig::corn_false_events_b}}
    \caption{
    Example of a filter result from the paper \cite{kowalczyk2022hardware} for a test sequence.
    In figure \ref{sfig::corn_all_events} the deleted events are shown in red and the passed events are shown in green.
    \ref{sfig::corn_false_events_a} shows only filtered events.
    \ref{sfig::corn_false_events_b} shows a close-up on the events removed due to the object's transition between neighbouring regions.
    }
    \label{fig::corn_all_events}
\end{figure}

Its greatest feature was the complete independence of operation of the areas into which the sensor matrix was divided. This could be seen by observing the performance of the algorithm for objects moving within the field of view of the event sensor.
Correct events were removed when an object moved between adjacent areas. This is because the area in question had not previously received any events, so its timestamp is too small for a new event to pass through.
Figure \ref{sfig::corn_all_events} shows the result of filtering for a test sequence containing a recording of falling corn kernels.
Events with a period of \SI{10}{\ms} are marked in the figure. The red colour indicates the events removed by the above filter, while the green colour indicates those passed to its output.
The filter parameters were as follows: filter length -- \SI{400}{\us}, area size -- 16 $\times$ 16, update factor -- 0.25.

Figure \ref{sfig::corn_false_events_a} shows only events filtered out by the same filter.
At the bottom of the graph, groups of rejected events are visible. They form below the lines that mark the edges between neighbouring regions. A close-up of these groups is shown in figure \ref{sfig::corn_false_events_b}. This is due to the phenomenon mentioned earlier - the timestamp of the new region is too small for new events to pass through for further processing.

\section{Interpolation-based algorithm}
\label{sec::interpolation}

The aim of this research was to reduce the effect described in section \ref{sec::basic_algorithm} by using the parameters of neighbouring areas.
The possibility of using algorithms to interpolate the timestamps of neighbouring areas to smooth the effect of objects passing between neighbouring areas was investigated. Several possibilities for combining these data were considered and their effectiveness analysed. The proposed methods used: timestamps of neighbouring areas, distances of events from neighbouring areas, frequencies of events in neighbouring areas.
Four algorithms were tested on the data: maximum timestamps, bilinear interpolation, bilinear interpolation with frequency weights, and distance-based interpolation with frequency weights.
A common part of all these algorithms is the estimation of the parameters of the areas into which the event sensor matrix has been divided.

The parameters of the algorithms developed are: sensor resolution, filter length -- the threshold that determines which events to filter out, scale -- the size of the areas into which the matrix is divided, update rate -- the rate at which the parameters of the areas are updated.
The transmitted events are processed one by one. Each event is first checked to see if it should be filtered out or sent for further processing. Then the parameters of the area to which the processed event belongs are updated.

The first parameter to be determined is the area's timestamp, which is the filtered timestamps of events in the area. Filtering makes it possible to retain some information about previous events in a given environment. As opposed to storing only the timestamp of the most recent event, this filtering makes it possible to achieve much greater resistance to high-intensity disturbances. The formula for calculating the timestamp is given in the equation \eqref{eq:timestamp}.

\begin{equation}
    Ts_{n+1} = Ts_n \cdot (1 - u) + Ts_e \cdot u
\label{eq:timestamp}
\end{equation}
where: \(Ts_{n}\) is the current timestamp of the region, \(Ts_{n+1}\) is the new timestamp of the region, \(u\) is the update factor, \(Ts_e\) is the timestamp of the processed event.

Another parameter to be determined was the event interval for the area under consideration. This should provide information on the frequency of events. The more frequently events occurred in an area, the greater its importance in interpolation. The formula describing the process of updating the event interval was presented in the equation \eqref{eq:interval}.

\begin{equation}
    I_{n+1} = I_n \cdot (1 - u) + (Ts_e - Ts_n) \cdot u
\label{eq:interval}
\end{equation}
where: \(I_n\) is the current interval of the area, \(I_{n+1}\) is the new interval of the area.

For each event processed, information was also stored in the area activity table. This was then used at defined intervals to globally modify all parameters that had not been updated at that time. This was done to reduce the number of discarded events when there had been no events for a long time in the area being analysed. It also made it possible to reduce the interval values for the same areas, so that their timestamps had less impact on the result of filtering with methods that take into account the frequency of events. This is done according to the formulas \eqref{eq:timestamp} and \eqref{eq:interval}, but the current time (timestamp of the last processed event) is used instead of the timestamp of the processed event.

\begin{figure}
    \centering
    \subfloat[]{\includegraphics[width=1.5in]{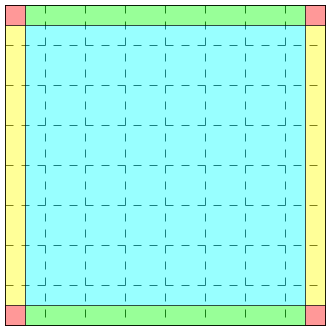}\label{sfig::Borders}}
    \hspace{.01in}
    \subfloat[]{\includegraphics[width=1.5in]{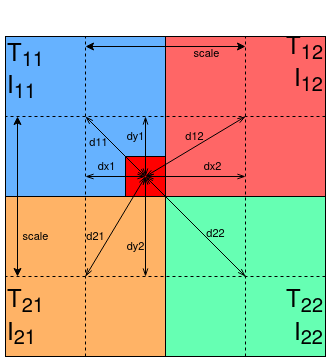}\label{sfig::Interpolation}}
    \caption{
    The \ref{sfig::Borders} shows the division of the matrix depending on the number of neighbouring regions for the event being processed, and the \ref{sfig::Interpolation} shows an example situation for the filtering algorithm.
    }
    \label{fig::BordersInterpolation}
\end{figure}

Depending on the position of the event to be processed on the matrix and the size of the regions into which it has been divided, a different number of neighbouring regions can be used for interpolation. This is illustrated in the figure \ref{sfig::Borders}. The dashed line shows the boundaries of the areas into which the sensor matrix has been divided. In the blue part, four neighbours are used for the interpolation. In the green and yellow parts two neighbours are used, horizontally and vertically respective. In the red part no interpolation is necessary. The interpolation algorithm therefore works slightly differently in each of these areas.

The operation of the four neighbourhood parameter interpolation methods analysed is described below. The source code for these methods in MATLAB can be found in the github repository\footnote{\url{https://github.com/vision-agh/DVS_FilterInterpolation}}.

Figure \ref{sfig::Interpolation} shows an example. The pixel for which the event was recorded is shown in red. The dashed lines mark the centres of the areas. The four adjacent areas are also visible. The parameters of these regions and the quantities calculated by the proposed methods are also marked.

\subsection*{Timestamps maximum (TM)}
\label{ssec::maximum}

The first method of combining data is to perform a simple max operation. This is described by the equation \eqref{eq::max}.
\begin{equation}
    T = max(T_{11}, T_{12}, T_{21}, T_{22})
\label{eq::max}
\end{equation}

It uses the timestamp of each adjacent area. It is denoted by \(T_{11}\), \(T_{12}\), \(T_{21}\) and \(T_{22}\). The maximum of these four plus the length of the filter is compared to the timestamp of the event being processed. If it is greater, the event is sent to the output.

\subsection*{Bilinear interpolation (BI)}
\label{ssec::bilinear}

The second method involves bilinear interpolation of area timestamps depending on the position of the pixel reporting the processed event. This is described by the equations \eqref{eq::bilinear_T12} and \eqref{eq::bilinear_T}.
\begin{equation}
\begin{split}
    T_1 &= \frac{T_{11} dx_2 + T_{12} dx_1}{scale}\quad
    T_2 = \frac{T_{21} dx_2 + T_{22} dx_1}{scale}
\end{split}
\label{eq::bilinear_T12}
\end{equation}
\begin{equation}
    T = \frac{T_1 dy_2 + T_2 dy_1}{scale}
\label{eq::bilinear_T}
\end{equation}
It also uses the timestamp of each neighbouring area. These are combined, taking into account the vertical and horizontal distance of the reporting pixel from the centres of the neighbouring areas. The result plus the length of the filter is the threshold for deciding whether an event is rejected or passed on for further processing.

\subsection*{Bilinear interpolation with frequency weights (BIF)}
\label{ssec::bilinearFrequency}

The next method is similar to the one previously discussed in the \ref{ssec::bilinear} subsection. However, it takes into account an additional parameter that gives an indirect indication of the number of events reported in neighbouring areas. This parameter is the average time between successive events. This is described by the equations \eqref{eq::bilinear_weights}.
\begin{equation}
\begin{split}
    T_1 &= \frac{T_{11} I_{12} dx_2 + T_{12} I_{11} dx_1}{I_{12} dx_2 + I_{11} dx_1}\\
    T_2 &= \frac{T_{21} I_{22} dx_2 + T_{22} I_{21} dx_1}{I_{22} dx_2 + I_{21} dx_1}\\
    T &= \frac{T_1 I_{21} I_{22} dy_2 + T_2 I_{11} I_{12} dy_1}{I_{21} I_{22} dy_2 + I_{11} I_{12} dy_1}
\end{split}
\label{eq::bilinear_weights}
\end{equation}
The more events reported for an area, the more weight it has in the threshold calculation. The calculation of the number of reported events is done by estimating the time between events for each area.
As with the previous methods, the result plus the length of the filter is the threshold that determines whether an event is rejected.

\subsection*{Distance-based interpolation with frequency weights (DIF)}

The last method calculates the threshold by taking into account the distance of the reporting pixel from the centres of the neighbouring areas and, as in the previous method, the frequency of events.
It uses the timestamp of each neighbouring area and its estimated interval between consecutive events. The assumption is that the weight of neighbouring areas should be proportional to the event frequency and inversely proportional to their distance from the event. Proportionality to frequency also implies inverse proportionality to the interval between events, and this quantity is calculated directly. This is described by the equations \eqref{eq::distance_C} and \eqref{eq::distance_T}.

\begin{equation}
\begin{split}
    C_{11} &= \frac{1}{I_{11} d_{11}}\quad
    C_{12} = \frac{1}{I_{12} d_{12}}\\
    C_{21} &= \frac{1}{I_{21} d_{21}}\quad
    C_{22} = \frac{1}{I_{22} d_{22}}
\end{split}
\label{eq::distance_C}
\end{equation}

\begin{equation}
    T = \frac{T_{11} C_{11} + T_{12} C_{12} + T_{21} C_{21} + T_{22} C_{22}}{C_{11} + C_{12} + C_{21} + C_{22}}
\label{eq::distance_T}
\end{equation}

The result of the above equation plus the length of the filter is the threshold that determines whether an event is rejected.

\section{Evaluation}
\label{sec::evaluation}
This chapter compares the proposed methods for combining neighbourhood parameters to filter event data. Several test datasets were used for this purpose.

\subsection*{Artificial noise}
\label{ssec:artificial_noise}
\begin{figure}[!t]
	\centering
	\includegraphics[width=3in]{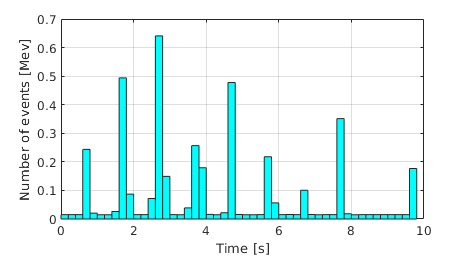}
	\caption{
     Example histogram of events for one of the falling corn sequences. Each bar represents the number of events in a \SI{0.2}{\s} time interval.
     }
	\label{fig::eventsHistogram}
\end{figure}

The original data was modified by adding artificial events to simulate interference in the event camera data stream.
The artificial noises were randomly generated events that did not take into account movement in the scene. They were then added to the test sequence with an indication that they were artificial data. 
In the original data, some of the events were also noise, but it was decided to make this additional modification in order to test the effectiveness of the algorithms for noise of different intensities. It also made it easier to assess how much of the noise was removed by the proposed methods and how much of the original signal is was preserved.
The first step in the analysis carried out was to assess the amount of noise in the recorded camera data sequences for a static scene. This can vary depending on the conditions of the observed scene (e.g. lighting) or the performance of the sensor.
To do this, histograms of the number of events in successive time intervals were calculated. An example of a histogram for a single sequence is shown in Figure \ref{fig::eventsHistogram}.

This makes it easy to distinguish the time intervals in which a falling object was recorded in the camera's field of view. The number of events is then significantly higher. To estimate the average number of events for a static scene, the highest and lowest bars were removed and the average value of the histogram patches for the remaining bars was calculated. In the example given, the 16 highest and 16 lowest values were removed and the result was 16297. Given that each interval was 0.2 s long, it can be estimated that the average noise rate for a static scene in this case was around \SI{0.09}{\Hz \per pix}. The estimate also shows that at least 19.87\% of the recorded events were disturbances. It should be noted that this is only the average amount of noise that is not related to motion in the scene.
Ideally, the events should only occur in the vicinity of the falling object (assuming this is the only moving object and the lighting conditions do not change). However, there are usually a large number of events in the event stream that are not related to object movement or changes in brightness. This can be seen in Figure \ref{fig::introduction} in the pre-filtration events (left).
The proposed filtering methods were used to filter several event camera datasets. The result of the filtering was evaluated in two ways.
The first way was to add artificially generated noise of varying intensity to the processed data and check the statistical results of the filtering. The quantities evaluated were the percentage of generated noise removed and the percentage of original events remaining after filtering. It should be noted, however, that some of the artificial noise may have fitted so well into the original data that it is not possible to distinguish them from the real events. It is therefore not possible to remove them all.
As discussed above, the original data also contains noise. The filtering algorithm should also be able to remove this as best it can. It is therefore expected that at least some of the original events will be removed. In addition, based on the estimation of the minimum proportion of distortions in the test set presented earlier, it is possible to assume a minimum proportion of the recorded events that should be removed during filtering.

The second way was to visually assess the result of the algorithm. This involved checking the transition times of moving objects between adjacent areas to see if events were incorrectly removed.
Three recordings of falling objects were used for the tests. The first two contained events recorded for falling grains of maize. These lasted 10 seconds. The third contained a recording of falling nuts lasting 6 seconds. Each of these recordings was analysed at different artificial noise intensities and for different filter lengths. For each set, the minimum interference estimation presented earlier was performed. The results of this estimation are given in the table \ref{tab:estimation}. The proportion of all events that are the result of the estimation performed is given, as well as the estimated number of interferences during one second of sensor operation.

\begin{table}[!t]
	\centering
	\begin{tabular}{| c | c | c |}
		\hline 
		{Dataset} & {\% noises} & {Noise rate[\si{\Hz \per pix}]} \\	
		\hline
		I     & 19.87 & 0.09 \\
        \hline
        II    & 18.47 & 0.08 \\
        \hline
        III   & 54.05 & 0.12 \\
		\hline
	\end{tabular}
	\caption{
        Estimation of the amount of noise in the datasets used
        }
	\label{tab:estimation}
\end{table}

\begin{figure}[!t]
	\centering
	\includegraphics[width=3in]{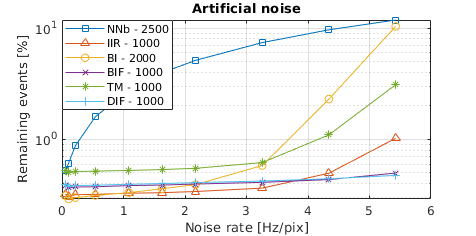}
	\caption{
    The amount of unremoved artificially generated artefacts by the methods analysed.
    }
	\label{fig::Comparison_RemainingArtificial_1000}
\end{figure}

\begin{figure}[!t]
	\centering
	\includegraphics[width=3in]{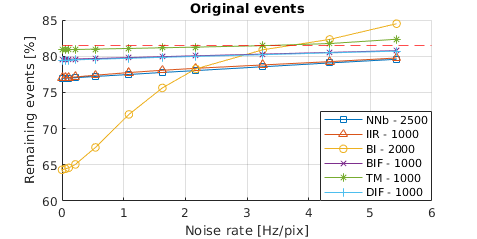}
	\caption{
    Number of original events left by the analysed methods.
    }
	\label{fig::Comparison_RemainingOriginal_1000}
\end{figure}

The performance of the proposed methods has been compared for the same filter length and different artificial noise intensities. The results for a test set are shown in the graphs \ref{fig::Comparison_RemainingArtificial_1000} and \ref{fig::Comparison_RemainingOriginal_1000}. The first shows the \% of artificial noise that has not been removed. The second shows the \% of the original noise that has been passed to the output. The red dashed line shows the estimate of the maximum number of original non-interference events based on the estimation described above. For all methods based on IIR filtering, the same parameters of the considered methods were used: filter length - \SI{1000}{\us}, area size - 16 $\times$ 16, update factor - 0.25.

Probably the most commonly used event data filtering algorithm, the NNb, has also been added to the comparison. This method checks whether at least one other event has been recorded in the vicinity of the processed pixel within the given time window. A time window of \SI{2500}{\us} was used, and the size of the neighbourhood was 3 $\times$ 3. Its length was chosen so as not to remove too much of the correct data.


From these graphs, several conclusions can be drawn about the algorithms analysed.
Firstly, the TM method 
passes the largest number of disturbances to the filter output. From a comparison with the estimation performed earlier, it can be seen that even for a small amount of artificial interference, this method removes a relatively small proportion of the interference. For a larger amount of noise, it removes even less than the estimation performed.
The efficiency of the IIR and BI methods also decreases significantly as the amount of noise increases.

Secondly, BI removes most of the artificial noise, but the difference is quite small at around 0.2\%. However, it also removes a very large proportion of the original data.
This is significantly larger than the estimated minimum noise content. It can therefore be concluded that a very large proportion of the correct data is also removed. It is also worth noting that the unfiltered part of the artificial noise increases more rapidly with the amount of noise than for the other methods.
An undesirable feature of this algorithm is that its effectiveness varies greatly with changes in noise intensity. In practical applications, this situation can occur very frequently (e.g. due to changes in lighting). The other proposed algorithms, in particular DIF and BIF, are very robust to noise variations.

Thirdly, the results of the two methods using weights based on event frequency are very similar. They remove the vast majority of the artificial noise and retain a fair proportion of the original events (including the estimated minimum number of noisy events). However, the first graph analysed shows that the BIF starts to lose efficiency sooner as the amount of noise increases.

Finally, the proposed methods have similar filtering efficiencies to the NNb method. However, it can be seen that the effectiveness of both frequency weighting methods scales better as the amount of interference increases. The NNb passes most of the artificially generated interference. This is due to the fact that as the interference intensity increases, the probability of two such events being very close together (both in time and space) increases, resulting in one of the events being passed to the filter output. The proposed methods are more robust in such situations.

\begin{figure}
    \centering
    \subfloat[]{\includegraphics[width=0.8in]{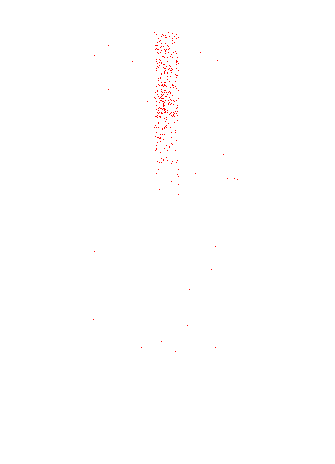}\label{sfig::corn_false_events_5_2_100}}
    \subfloat[]{\includegraphics[width=0.8in]{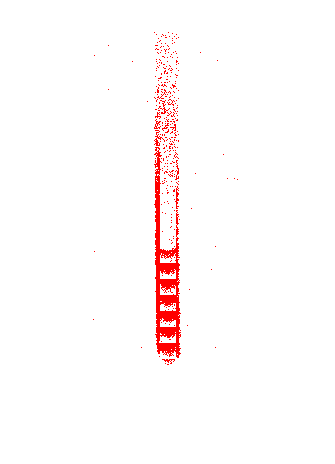}\label{sfig::corn_false_events_5_0_100}}
    \subfloat[]{\includegraphics[width=0.8in]{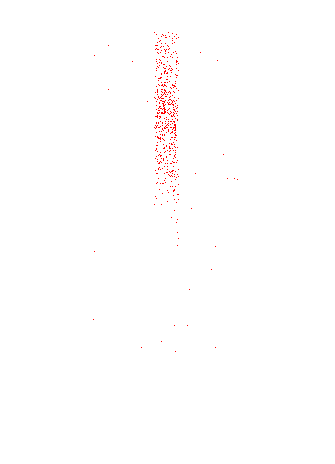}\label{sfig::corn_false_events_5_1_100}}
    \subfloat[]{\includegraphics[width=0.8in]{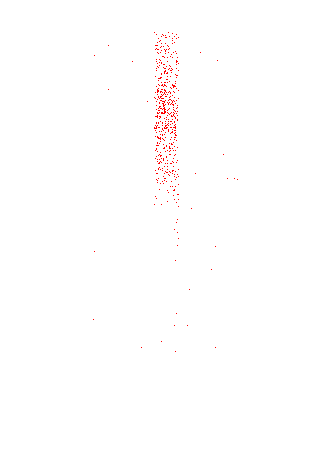}\label{sfig::corn_false_events_5_3_100}}
    \caption{
    Filtered events for the falling corn record. \ref{sfig::corn_false_events_5_2_100} refers to TM, \ref{sfig::corn_false_events_5_0_100} to BI, \ref{sfig::corn_false_events_5_1_100} to BIF, and \ref{sfig::corn_false_events_5_3_100} to DIF.
    }
    \label{fig::corn_false_events_5_100}
\end{figure}

The final element of the evaluation was to check whether the proposed algorithms achieve the stated goal of a small fraction of events removed when a moving object passes between adjacent areas. For this purpose, a performance comparison was made similar to the result of the original filtering algorithm \ref{sfig::corn_false_events_a} presented earlier. These are shown in figure \ref{fig::corn_false_events_5_100}. They show the result of each of the methods analysed. The same parameters were used for each of them: filter length - \SI{400}{\us}, domain size - 16 $\times$ 16, update factor - 0.25.
They were taken on the same moment of the same test dataset.
Several conclusions can be drawn.

Firstly, Method II removes a very large proportion of correct events. This shows that BI has a negative impact on the efficiency of the algorithm. This is due to the fact that active and inactive areas affect events in the area between their centres in the same way. As a result, events that would have been passed to the output in the original version of the algorithm may be removed after using this method. The use of this method is therefore counterproductive.

Secondly, the result of method I shows the fewest events. It fulfils the stated requirement, i.e. it does not remove events when objects pass between neighbouring areas. On the other hand, it also marks far fewer elements that should be marked as noise. This means that it removes the least number of noisy events of the proposed methods. It is therefore less effective than the other proposed methods in terms of noise removal.

Thirdly, both method III and method IV achieve the stated objective of no removed interference when a moving object passes between adjacent areas. The result of these methods is very similar and each produces similar results. They do a good job of removing interference without removing too many valid events. For method III there are 580 events highlighted in the window shown, while for method IV there are 581. The events removed are not exactly the same. If you look closely at the images \ref{sfig::corn_false_events_5_1_100} and \ref{sfig::corn_false_events_5_3_100} you can see differences, but they are very small. Therefore, it can be concluded that both methods do a good job of removing noise and do not remove an excessive number of relevant events.

A requirement for the proposed filtering methods was also a low memory requirement. Therefore, it was calculated how much of the coefficients the proposed methods need to store in memory for a sensor with a resolution of 1280 $\times$ 720. Methods I and II require \SI{14.85}{KB} of data to be stored in memory, while methods III and IV require \SI{29.7}{KB}. In comparison, methods that require a timestamp to be stored for each pixel require approximately \SI{3.67}{MB} of data to be stored in memory.

\subsection*{Recorded noise}
\label{ssec:recorded_noise}

The methodology for evaluation with recorded noise data was inspired by paper \cite{guo2022low}. The authors proposed to record noise data from a DVS sensor, and then insert in into a clean event stream. This methodology make it easy to distinguish the noise and signal data at the output of the evaluated filter. Moreover, they have also proposed a way to evaluate filer efficiency without using a single discrimination threshold. Instead, they have used a Receiver Operating Characteristic (ROC). It plots True Positive Rate (TPR) and False Positive Rate (FPR) across different discrimination thresholds.
A measure of filtration quality is the area under the curve (AUC) calculated for the ROC plot. A higher AUC value indicates better filtration quality.

In order to apply the methodology described, several sequences containing only noise were recorded using the event camera \textit{Prophesee EVK1 - Gen4 HD}.
This involved positioning the camera so that there were no moving objects within its field of view and no changes in illumination.
Several sets of data were then recorded for different levels of illumination. To further increase the range of recorded noise, the camera operating parameters responsible for sensitivity to changes in incident light intensity were changed.
To test the effectiveness of the filtration in different environments and situations, it was decided to record a few test sequences.
Finally, four test sequences were used. The first was a previously used sequence of falling maize grains. It is designed to simulate a situation where the system's objective is to count falling objects. Due to the small size of the objects and their high speed, this collection is very sparse.
The second sequence contains a recording of people dancing in an enclosed space. It simulates the situation when a stationary camera observes a scene with a lot of movement.
The third sequence contains footage of people in a closed room using a moving event camera.
The fourth sequence contains data recorded while the camera is moving in an urban environment.

These were filtered to remove as much noise as possible. The previously recorded noise was then added to the filtered data. This produced a set of sequences containing known noise and the actual data.
The generated sequences were then filtered using the proposed filtering methods. The AUC coefficients for these methods and the different test sequences are shown in the Table \ref{tab::auc}. In the results shown, the recorded noise with the intensity of around \SI{1.82}{\Hz\per pix}, was used.
\begin{table}[!t]
	\centering
	\begin{tabular}{| c | c | c | c | c |}
		\hline 
            {Algorithm} & {I} & {II} & {III} & {IV} \\	
		\hline
            NNb &   0.958   & 0.924 & 0.867 & 0.779 \\
            \hline
            IIR &   0.995   & 0.956 & 0.899 & 0.838 \\
            \hline
            BI &    0.998   & 0.963 & 0.906 & 0.848 \\
            \hline
            BIF &   0.999   & 0.961 & 0.903 & 0.841 \\
            \hline
            TM &    0.998   & 0.950 & 0.881 & 0.816 \\
            \hline
            DIF &   0.999   & 0.960 & 0.901 & 0.842 \\
		\hline
	\end{tabular}
	\caption{
        AUC values for the proposed filtration methods and different test scenarios
        }
	\label{tab::auc}
\end{table}
From the results presented, it can be noticed that the proposed methods perform very well in situations where a stationary camera captures motion in the observed scene. They perform significantly less well when processing events from a moving sensor.




\section{Conclusion}
\label{sec::conclusion}
The aim of the work was to propose a method of event data filtering that would have high performance but would not require a large amount of data to be stored in memory. 
We proposed 4 methods for event data filtering based on the method described in the paper \cite{kowalczyk2022hardware}. It requires a small amount of memory, but a significant proportion of valid events are removed when an object moving in the camera's field of view passes between adjacent areas into which the matrix has been divided.
The proposed algorithms consisted of interpolating data from adjacent areas so that the passage of objects between the areas would not result in the deletion of correct events.
The first method consisted of finding the maximum of the timestamps of adjacent regions, the second of their bilinear interpolation, the third additionally took into account the weights of these areas based on the event frequency, and the last calculated the event distance to the centres of the neighbouring areas and the weights based on the event frequency.
Their description is given in chapter \ref{sec::interpolation}. Chapter \ref{sec::evaluation} compares their performance and memory requirements for a 1280 $\times$ 720 resolution sensor.

Based on the results presented, the DIF algorithm seems to perform best. However, its advantage over the BIF is small. From the evaluation carried out, it can also be concluded that the desired objective, i.e. high filtering efficiency with low memory requirements, has been achieved. In order to evaluate this, a comparison was also made between the proposed methods and probably the most commonly used method for filtering event data, i.e. NNb.
The proposed methods perform better when a stationary camera observes moving objects than when the sensor itself is moving.

As part of further work, it is planned to implement the selected algorithm in an FPGA chip. This will allow a significant acceleration of the calculations performed. This implementation will be possible due to the low memory consumption of the proposed method.

\section{Acknowledgements}
\label{sec::acknowledgements}
The work presented in this paper was supported by the National Science Centre project no. 2021/41/N/ST6/03915 entitled ``Acceleration of processing event-based visual data with the use of heterogeneous, reprogrammable computing devices'' (first author) and the programme ``Excellence initiative – research university'' for the AGH University of Science and Technology.

{\small
\bibliographystyle{ieee_fullname}
\bibliography{egbib}
}

\end{document}